\documentclass[10pt,twocolumn,letterpaper]{article}

\usepackage{cvpr}              

\usepackage{graphicx}
\usepackage{amsmath}
\usepackage{amssymb}
\usepackage{booktabs}
\usepackage{multirow}
\usepackage{subcaption}
\usepackage[accsupp]{axessibility}  

\newcommand{\VS}{\textit{vs}.\xspace}

\usepackage[pagebackref,breaklinks,colorlinks]{hyperref}

\usepackage[capitalize]{cleveref}
\crefname{section}{Sec.}{Secs.}
\Crefname{section}{Section}{Sections}
\Crefname{table}{Table}{Tables}
\crefname{table}{Tab.}{Tabs.}

\def\confName{CVPR}
\def\confYear{2023}

\begin{document}

\title{HNeRV: A Hybrid Neural Representation for Videos}

\author{Hao Chen$^{1}$\quad Matthew Gwilliam$^{1}$ \quad Ser-Nam Lim$^{2}$\quad Abhinav Shrivastava$^{1}$\\[0.5em]
$^{1}$University of Maryland, College Park \quad\quad $^{2}$Meta AI \quad\quad \\
{\tt\small \{chenh,mgwillia,abhinav\}@cs.umd.edu, sernamlim@meta.com} \\
{\tt\small ~\url{https://haochen-rye.github.io/HNeRV/}}
}
\maketitle

\begin{abstract}
    Implicit neural representations store videos as neural networks and have performed well for various vision tasks such as video compression and denoising. 
    With frame index or positional index as input, implicit representations (NeRV, E-NeRV, \etc) reconstruct video frames from  \textit{fixed} and \textit{content-agnostic} embeddings. 
    Such embedding largely limits the regression capacity and internal generalization for video interpolation. 
    In this paper, we propose a Hybrid Neural Representation for Videos (\textbf{HNeRV}), where a \textit{learnable} encoder generates \textit{content-adaptive} embeddings, which act as the decoder input. 
    Besides the input embedding, we introduce HNeRV blocks, which ensure model parameters are evenly distributed across the entire network, such that higher layers (layers near the output) can have more capacity to store high-resolution content and video details. 
    With content-adaptive embeddings and re-designed architecture, HNeRV outperforms implicit methods in video regression tasks for both reconstruction quality ($+4.7$ PSNR) and convergence speed ($16\times$ faster), and shows better internal generalization. 
    As a simple and efficient video representation, HNeRV also shows decoding advantages for speed, flexibility, and deployment, compared to traditional codecs~(H.264, H.265) and learning-based compression methods. 
    Finally, we explore the effectiveness of HNeRV on downstream tasks such as video compression and video inpainting.
\end{abstract}


\section{Introduction}
\begin{figure}[t!]
    \centering
    \includegraphics[width=.98\linewidth]{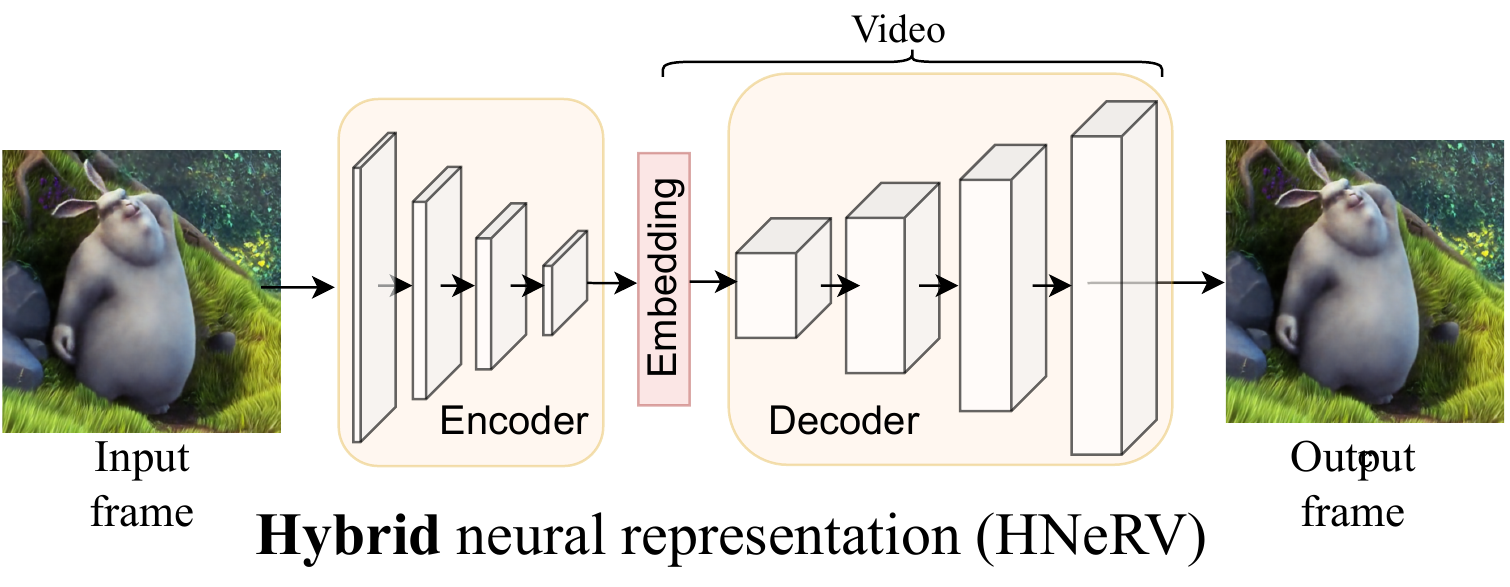} 
    \includegraphics[width=.9\linewidth]{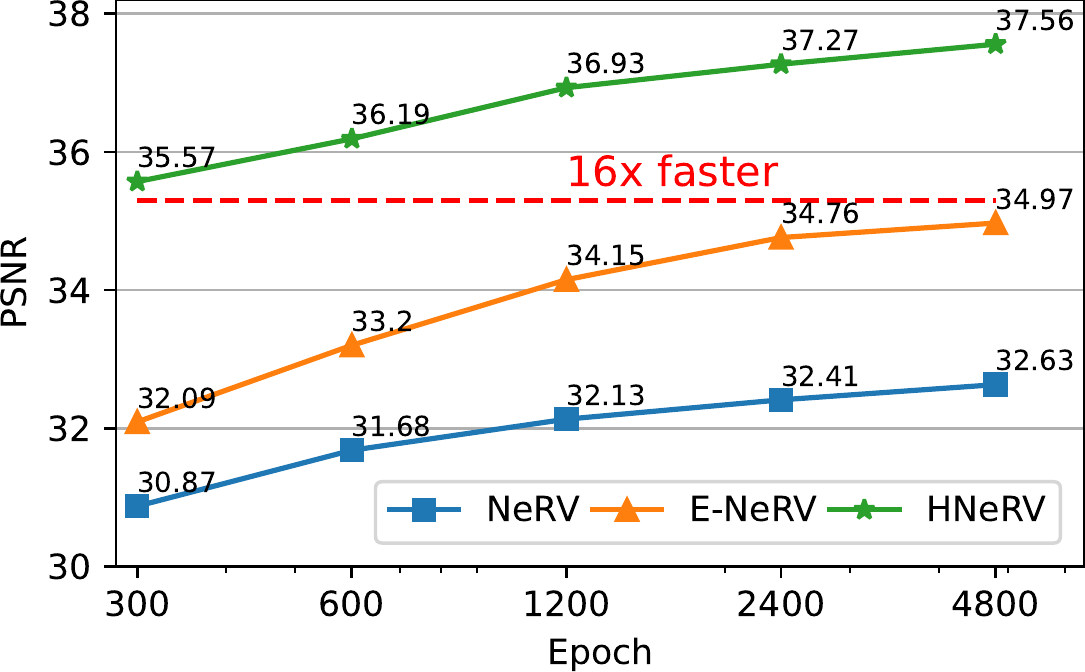}   
    \caption{
    \textbf{Top:} hybrid neural representation with learnable and content-adaptive embedding (ours).
    \textbf{Bottom:} video regression for hybrid and implicit neural representations.    
  }
    \label{fig:hnerv-teaser}
\end{figure}

Given the massive amount of videos generated every day, storing and transferring them efficiently is a key task in computer vision and video processing.
Even for modern storage systems, the space requirements of raw video data can be overwhelming. 
Despite storage becoming cheaper, network speeds and I/O processing remain a bottleneck and make transferring and processing videos expensive.

Traditional video codecs, such as H.264~\cite{H264} and HEVC~\cite{hevc}, rely on a manually-designed encoder and decoder based on discrete cosine transform~\cite{dct}. 
With the success of deep learning, many attempts~\cite{lu2019dvc,rippel2021elfvc,Agustsson_2020_CVPR,Djelouah_2019_ICCV,Habibian_2019_ICCV,liu2019neural,9247134,Rippel_2019_ICCV,Wu_2018_ECCV} have been made to replace certain components of existing compression pipelines with neural networks. 
Although these learning-based compression methods show high potential in terms of rate-distortion performance, they suffer from complex pipelines and expensive computation, not just to train, but also to encode and decode.

To address the complex pipelines and heavy computation, implicit neural representations~\cite{pmlr-v97-rahaman19a,sitzmann2020implicit,schwarz2021graf,Chen_2019_CVPR,Park_2019_CVPR} have become popular due to their simplicity, compactness, and efficiency. 
These methods show great potential for visual data compression, such as COIN~\cite{dupont2021coin} for image compression, and NeRV~\cite{chen2021nerv} for video compression. 
By representing videos as neural networks, video compression problems can be converted to model compression problems, which greatly simplifies the encoding and decoding pipeline. 

Implicit representation methods for video compression present a major trade-off: they embrace simplicity at the expense of generalizability.
Given a frame index $t$ as input, NeRV~\cite{chen2021nerv} uses a fixed position encoding function and a learnable decoder to reconstruct video frames from temporal embeddings.
Another implicit representation, E-NeRV~\cite{li2022enerv}, takes a temporal embedding and spatial embedding to reconstruct video frames.
Since the embeddings of NeRV and E-NeRV are based on spatial and/or temporal information only; without connection to the actual content of frames, they are content-agnostic. 
For decoding, NeRV-like models compute these embeddings using frame index alone, without access to the original frame.
This is quite elegant for video compression, since instead of storing many frame embeddings, one would only need to store model weights and basic metadata (\eg, number of frames). 

However, this comes with some major disadvantages.
Firstly, since embeddings are content-agnostic, and due to how the temporal embeddings are computed, there is no way to meaningfully interpolate between frames.
Secondly, and more importantly, the positional embedding used by the fully-implicit models provides no visual prior and limits the regression capacity, since all the information needs to be learned by and stored in the video decoder.

In this paper, we propose a learnable encoder as a key component of hybrid neural representation for videos~(HNeRV, Figure~\ref{fig:hnerv-teaser} (top).
Our proposed neural representation is a hybrid between implicit (network-centric) and explicit (embedding-centric) approaches since it stores videos in two parts: the tiny content-adaptive frame embeddings and a learned neural decoder.
Besides the issue of content-agnostic embedding, prior work such as NeRV also suffers from an imbalance in the distribution of model parameters. 
In these decoders, later layers (closer to the output image) have much fewer parameters than earlier layers (closer to the embedding).
This hinders NeRV's ability to effectively reconstruct massive video content while preserving frame details.
To rectify this, we introduce the HNeRV block, which increases kernel sizes and channel widths at later stages.
With HNeRV blocks, we can build video decoders with parameters that are more evenly distributed over the entire network.
As a hybrid method, HNeRV improves reconstruction quality for video regression and boosts the convergence speed by up to $16\times$ compared to implicit methods, shown in Figure~\ref{fig:hnerv-teaser} (bottom).
With content-adaptive embeddings, HNeRV also shows much better internal generalization (ability to encode and decode frames from the video that were not seen during training), and we verify this by frame interpolation results in Section~\ref{subsec:main-results}.

HNeRV only requires a network forward operation for video decoding, which offers great advantages over traditional codecs and prior deep learning approaches in terms of speed, flexibility, and ease of deployment.
Additionally, most other video compression methods are auto-regressive and there is a high dependency on the sequential order of video frames.
In contrast, there is no dependency on the sequential order of frames for HNeRV, which means it can randomly access frames efficiently to decode frames in parallel.
Such simplicity and parallelism make HNeRV a good codec for further speedups, like a special neural processing unit (NPU) chip, or parallel decoding with huge batches.

HNeRV is still viable for video compression, while also showing promising performance for video restoration tasks.
We design our encoder such that it can also be compressed; additionally, our HNeRV decoder blocks perform well in the model compression regime, such that HNeRV is competitive with state-of-the-art methods.
We posit that neural representation can be robust to distortion in pixel space and therefore restore videos which have undergone distortions.
We verify this observation on the video inpainting task.

In summary, we propose a hybrid neural representation for videos. 
With content-adaptive embedding and re-designed architecture, HNeRV shows much better video regression performance over implicit methods, in reconstruction quality ($+4.7$ PSNR), convergence speed ($16\times$ faster), and internal generalization.
As an efficient video codec, HNeRV is easy to deploy, and is simple, fast, and flexible during video decoding.
Finally, HNeRV shows good performance over downstream tasks like video compression and video inpainting.


\section{Related Work}

\begin{figure*}[t!]
    \centering
    \includegraphics[width=.85\linewidth]{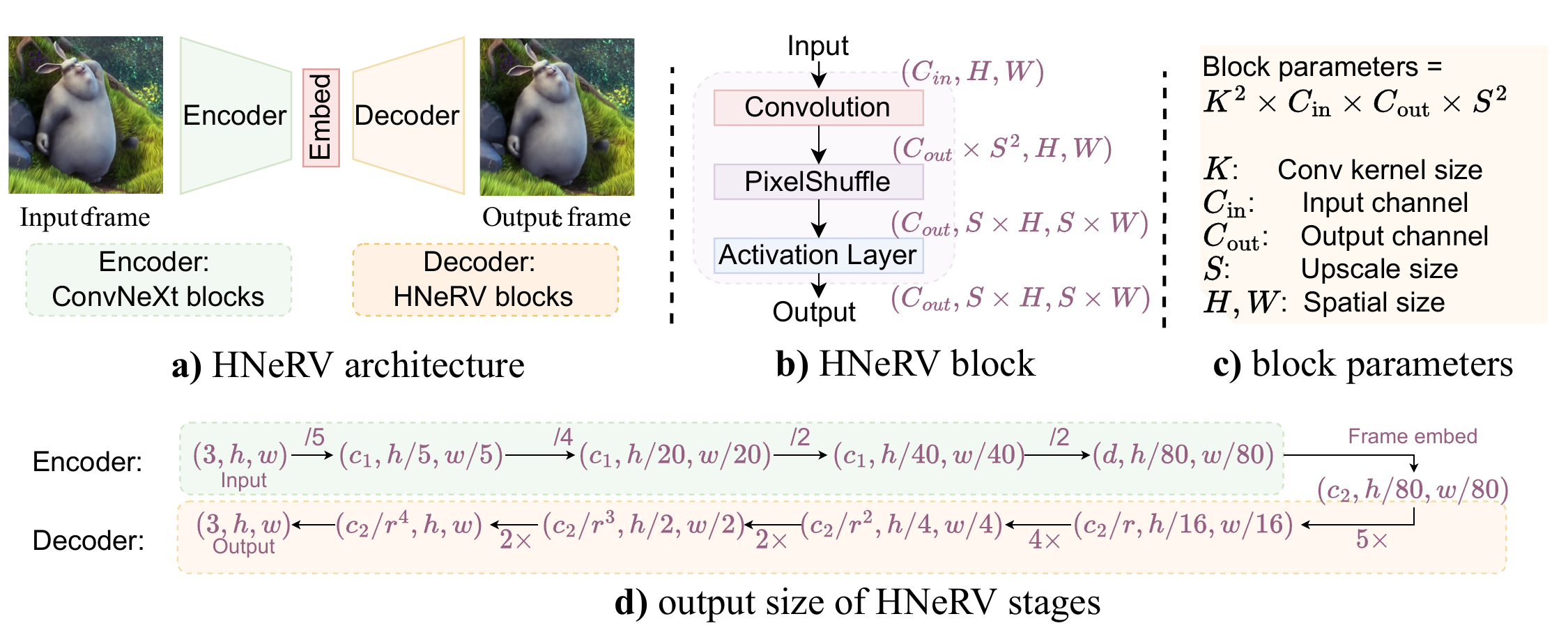} 
    \caption{
    \textbf{a)} HNeRV uses ConvNeXt blocks to encode frames as tiny embeddings, which are decoded by HNeRV blocks.
    \textbf{b)} HNeRV blocks consist of three layers: convolution, PixelShuffle, and activation (with input/output size illustrated).
    \textbf{c)} We demonstrate how to compute parameters for a given HNeRV block.
    \textbf{d)} Output size of each stage with strides 5,4,2,2.}
    \label{fig:hnerv-archi}
    \vspace{-.5em}
\end{figure*}

\noindent\textbf{Neural Representation.}
Implicit representations fit to each individual test signal~\cite{mehta2021modulated} where the model is regressed to a given image, scene, or video.
Most implicit neural representations are coordinate-based.
These coordinate-based implicit representations are used in image reconstruction~\cite{tancik2020fourier,sitzmann2020implicit}, shape regression~\cite{Chen_2019_CVPR,Park_2019_CVPR}, and 3D view synthesis~\cite{mildenhall2020nerf,schwarz2021graf}.
NeRV~\cite{chen2021nerv} instead proposes an image-wise implicit representation, which takes frame indices as inputs and leverages neural representation for fast and accurate video compression. 
Relying only on index, and not coordinates, speeds up the encoding and decoding process compared to coordinate-based (pixel-wise) methods. 
Based on NeRV, CNeRV~\cite{hao2022cnerv}, DNeRV~\cite{he2023dnerv}, E-NeRV~\cite{li2022enerv}, and NIRVANA~\cite{maiya2023nirvana} further improves the video regression performance.
Traditional autoencoders would not be considered implicit representations since most information is stored in their large image-specific embeddings.
Nevertheless, they are a form of neural representation, and HNeRV borrows the general concept for its encoder from standard $U$-shaped autoencoders~\cite{dct,autoencoder_08,kingma2014autoencoding,NIPS2016_eb86d510}.
HNeRV keeps the embedding intentionally tiny and compact, so as to keep most of the representation implicit (stored in the decoder).

\noindent\textbf{Video Compression.}
Traditional video compression methods such as MPEG~\cite{mpeg}, H.264~\cite{wiegand2003overview}, and H.265~\cite{hevc} achieve good reconstruction results with decent decompression speeds.
Recently, deep learning techniques have been proposed for video compression.
While these approaches focus on replacing the entire compression pipeline, they each borrow principles from the traditional handcrafted approaches.
Some have framed the problem primarily as image compression and interpolation~\cite{Wu_2018_ECCV,Djelouah_2019_ICCV}, or attempt to solve this task with image compression via autoencoders~\cite{Habibian_2019_ICCV}, or focus purely on interpolation for the sake of compression~\cite{liu2020conditional}.
Others essentially reformulate traditional video compression pipelines using deep learning tools~\cite{Rippel_2019_ICCV,liu2019neural,Agustsson_2020_CVPR}, at varying levels of complexity.
Recent approaches have focused on tackling the computational inefficiencies of existing art, including by fine-tuning traditional codecs~\cite{khani2021efficient}, and by optimizing pieces of the compression pipeline~\cite{rippel2021elfvc}.
The approach which inspired much of this work, NeRV, responds to these same inefficiencies by proposing a specialized architecture for video memorization~\cite{chen2021nerv}.
Once video is represented as a neural network, the video compression problem can be converted to a model compression problem and achieve good bit-distortion performance.
With learnable embeddings and re-designed decoder blocks, HNeRV improves the video regression capacity and convergence speed, while video compression is still viable by model compression.

\noindent\textbf{Model Compression.}
NeRV formulated video compression as model compression~\cite{chen2021nerv}, which is a diverse area.
In this paper we apply only a small subset of possible methods.
We use weight pruning~\cite{DBLP:journals/corr/HanPTD15} and weight quantization~\cite{DBLP:journals/corr/GuptaAGN15,DBLP:journals/corr/abs-1712-05877,krishnamoorthi2018quantizing}.
We also use entropy encoding for lossless compression after pruning and quantization~\cite{huffman1952method,han2015deep}.
Note that many other model compression methods can be leveraged to further reduce the size and video neural representation can always benefit from developments in the model compression area.

\noindent\textbf{Video Inpainting \& Internal Learning.} 
Video inpainting is typically framed as some combination of object removal and attempting to recreate missing regions of images.
Whereas some methods rely on priors from training on large datasets~\cite{wang2021image}, ours has more in common with a recent zero-shot fully-internal approach~\cite{ouyang2021video}.
We define ``Internal learning'' in terms of exploiting recurrence of information within a single domain, like within an image~\cite{Shocher_2018_CVPR}) or within an video ~\cite{zhang2019internal}.
It can be thought of as a sort of DIP-for-video, a line of work that was started for images with DIP~\cite{ulyanov2018deep} and extended for video by double-DIP~\cite{Gandelsman_2019_CVPR}.
Other methods have embraced this paradigm partially, learning some priors from large external datasets, before learning video-specific priors via internal learning~\cite{wang2021image}.

\section{Method}

We first give an overview of HNeRV (\cref{sec:hnerv-overview}).
We explain what makes HNeRV a ``hybrid'' representation, and the advantages that this offers.
We provide architectural details, loss functions, and explain how we compute the size of our video representation.
We then give particulars necessary for utilizing HNeRV on downstream tasks (\cref{sec:downstream-task}).
These include vector quantization for video compression and reconstruction loss for inpainting.

\subsection{HNeRV overview}
\label{sec:hnerv-overview}

Compared with a simple auto-encoder which uses one model for \textit{all} videos, and has \textit{large} frame embeddings, HNeRV (Figure~\ref{fig:hnerv-archi} (a)) fits a model on \textit{each} video, and has \textit{tiny} frame embeddings. 
We utilize ConvNeXt blocks~\cite{liu2022convnet} to build the encoder, and propose novel HNeRV blocks (Figure~\ref{fig:hnerv-archi} (b)) to build the decoder.

\noindent\textbf{Hybrid Neural Representation.} 
Compact video representations can be divided into two parts: explicit methods and implicit methods.
Explicit methods use an autoencoder to encode and decode all videos, and store content \textit{explicitly} as a latent embedding.
Given a \textit{video-specific embedding} as input, the decoder can reconstruct the video.
Implicit methods use only a learnable decoder to represent the video.
Given fixed frame index as input, the \textit{video-specific decoder} can reconstruct the video.
With content-adaptive embedding as input, explicit representation shows better generalization and compression performance, while implicit representations have a much simpler encoding/decoding pipeline and a high potential for compression (benefits from model compression techniques) and other downstream tasks (\eg efficient video dataloader, video denoising, inpainting).
In this paper, we propose a hybrid neural representation to combine the advantages of both explicit and implicit methods.
Similar to implicit representation, we use a learnable decoder to model video separately and store most content implicitly in the video-specific decoder.
To achieve better reconstruction, we use a learnable embedding as input and store information explicitly in these frame-specific embeddings, which is similar to explicit methods.
Therefore, we can use any powerful encoder to generate tiny content-adaptive embeddings to boost the performance of implicit representation.
Since these embeddings are quite small (\eg a $128$-d vector for a $640\times1280$ frame), our hybrid neural representation is as compact as implicit methods, but with stronger capacity, faster convergence, and better internal generalization, while keeping the full potential for downstream tasks.

\noindent\textbf{Model Architecture.} 
Similar to a NeRV block, an HNeRV block consists of three layers: convolution layer, pixelshuffle layer, and activation layer.
Within each block, only the convolution layer has learnable parameters (Figure~\ref{fig:hnerv-archi} (b)).
Illustrated in Table~\ref{tab:blk-comparison}, a NeRV block uses fixed kernel sizes for all stages $K=3$, and reduces channel width by 2, C$_\text{out}=C_\text{in}/2$.
Therefore, for blocks at later stages, the parameters are quite few and may not be strong enough to store video content at high resolution.
In contrast, we increase the kernel size and channel width for later HNeRV blocks, where $K$ increases from 1 (stage 1), to 3 (stage 2), to $K_\text{max}$ ($5$, \etc for later stages), and we decrease channel width by a reduction factor $r$ ($1.2$, \etc).
With kernel size $1$, the first block has much fewer parameters; with larger kernel size and wider channels, HNeRV blocks at later stages are much stronger; and we therefore get a more even distribution of model parameters across layers.
We list output sizes of various stages in Figure~\ref{fig:hnerv-archi} (d), 
with embedding dimension $d$ and channel reduction $r$.
Each stage has one block, and we use a $1\times1$ convolution layer to get low-dimension frame embeddings (channel width from $c_1$ to $d$), and a $3\times3$ convolution layer for final image predictions (channel width from $c_2/r^4$ to $3$).

\begin{table}[t!]
    \vspace{-0.7em}
    \centering
    \caption{\textbf{HNeRV block \VS NeRV block}. 
    $k$ is kernel size for each stage, $C_\text{out}$ and $C_\text{in}$ are output/input channels for each block.
    We decrease parameters via a small $k = 1$ for first block, and increase parameters for later layers with a larger $k$ and wider channels.}
    \label{tab:blk-comparison}    
    \vspace{-0.5em}
\resizebox{.8\linewidth}{!}{
    \begin{tabular}{l|c}
    \toprule
         NeRV blocks & $k=3, \quad C_\text{out}=C_\text{in}/2$  \\
         \midrule
    HNeRV blocks  &  $k=1,3,...,K_\text{max}, \quad C_\text{out}=C_\text{in}/r$ \\
         \bottomrule
    \end{tabular}
    }
    \vspace{-0.5em}
\end{table}

\noindent\textbf{Loss Functions.}
Since HNeRV attempts to reconstruct video with high fidelity, we use the loss objective
$L = \text{Loss}(x, p)$,
where $x$ is the input frame, $p$ is the HNeRV prediction, and `Loss' is any reconstruction loss function like L2, L1, or SSIM loss.

\noindent\textbf{Total size.}
As a hybrid neural representation, we include both frame embedding and decoder parameters to compute the total size of our video representation:
$\text{TotalSize} = \text{EmbedSize} + \text{DecoderSize}$.

\subsection{Downstream tasks}
\label{sec:downstream-task}

\noindent\textbf{Video Compression.}
We leverage both model compression and embedding quantization for video compression. 
Similar to NeRV, we apply global unstrutured pruning, model quantization, and weight entropy encoding for model compression. 
\textit{Note that unlike NeRV, which only stores non-zero parameters but no mechanism for mapping stored weights to their location in the network (and is thus impractical and unfair for compression comparison) we use entropy encoding to store the sparse weights (details can be found in the appendix). 
}

For quantization of a vector $\mu$, we linearly map every element to the closest integer,
\begin{equation}
  \begin{aligned}
    \mu_i = \text{Round}\left(\frac{\mu_i - \mu_\text{min}}{\text{scale}}\right) &* \text{scale} + \mu_\text{min} , 
    \text{where } \\
    \text{scale} = & \frac{\mu_\text{max} - \mu_\text{min}}{2^{b} - 1} ,
    \label{equa:quant}
  \end{aligned}
\end{equation}
$\mu_{i}$ is vector element, `Round' is a function that rounds to the closest integer, $b$ is the quantization bit length, $\mu_\text{max}$ and $\mu_\text{min}$ are the max and min value of vector $\mu$, and `scale' is the scaling factor.

\noindent\textbf{Video Inpainting.} 
For partially distorted video, we only compute loss for non-masked pixels, 
\begin{equation}
    L_\text{inpainting} = (1 - M) * \text{Loss}(x, p)
    \label{equa:inpaint-loss}
\end{equation}
where $M$ is the mask matrix where distorted pixels are $1$ and other are $0$. 
For inpainting output, following IIVI~\cite{ouyang2021video}, we fill the masked region with HNeRV's output.

\begin{table*}[t!]
\begin{minipage}{0.49\linewidth}
\centering
\caption{Video regression with different \textbf{sizes}}
\label{tab:bunny-size}
\vspace{-0.5em}
    \begin{tabular}{@{}l|cccc|c@{}}
    \toprule
    Size   & 0.35M & 0.75M & 1.5M  & 3M    & avg. \\
    \midrule
    NeRV   & 26.99 & 28.46 & 30.87 & 33.21 & 29.88                   \\
    E-NeRV & 27.84 & 30.95 & 32.09 & 36.72 & 31.90                   \\
    HNeRV  & \textbf{30.15} & \textbf{32.81} & \textbf{35.57} &\textbf{ 37.43 }& \textbf{33.90}                  \\
    \bottomrule
    \end{tabular}
\end{minipage} 
\hfill
\begin{minipage}{0.49\linewidth}
\caption{Video regression with different \textbf{epochs}} 
\label{tab:bunny-epoch}
\vspace{-0.5em}
\resizebox{.98\linewidth}{!}{
    \begin{tabular}{@{}l|cccccc@{}}
    \toprule
    Epoch  & 300   & 600   & 1200  & 1800 & 2400 & 3600    \\
    \midrule
NeRV &  	28.46 &	29.15 &	29.57 &	29.73 &	29.77 &	29.86 \\
E-NeRV &	30.95 &	32.07 &	32.79 &	33.1 &	33.36 &	33.67 \\
HNeRV &	\textbf{32.81} &	\textbf{33.89} &	\textbf{34.51} &	\textbf{34.73} &	\textbf{34.88} &	\textbf{35.03}  \\ 
    \bottomrule
    \end{tabular}
}
\end{minipage} 
\end{table*}

\begin{table*}[t!]
\centering
\caption{Video regression at resolution \textbf{960}$\times$\textbf{1920}, PSNR$\uparrow$ reported}
\label{tab:video-960p}
\vspace{-0.5em}
\begin{tabular}{@{}l|ccccccccccc|c@{}}
\toprule
Video & beauty & swan & bmx & bosph & dance & camel & bee & jockey & ready & shake & yach & avg.   \\
\midrule
NeRV    & 33.25  & 28.48          & 27.86          & 33.22    & 26.45           & 24.81      & 37.26    & 31.74  & 24.84         & 33.08     & 28.30     & 29.94 \\
HNeRV   & \textbf{33.58}  & \textbf{30.35}          & \textbf{29.98}          & \textbf{34.73}    & \textbf{30.45  }         & \textbf{26.71  }    & \textbf{38.96}    & \textbf{32.04}  & \textbf{25.74  }       & \textbf{34.57 }    & \textbf{29.26}     & \textbf{31.49} \\
\bottomrule
\end{tabular}

\caption{Video regression at resolution \textbf{480}$\times$\textbf{960}, PSNR$\uparrow$ reported}
\label{tab:video-480p}
\begin{tabular}{@{}l|ccccccccccc|c@{}}
\toprule
Video & beauty & swan & bmx & bosph & dance & camel & bee & jockey & ready & shake & yach & avg.   \\
\midrule
NeRV    & 36.27  & 29.75          & 28.81          & 35.07    & 29.47           & 26.75      & 40.76    & 32.58  & 25.81         & 35.33     & 30.11     & 31.88 \\
HNeRV   & \textbf{36.91}  & \textbf{31.92}          & \textbf{31.27}          & \textbf{36.95}    & \textbf{33.85}           & \textbf{28.85}      & 42.05    & \textbf{33.33}  & \textbf{27.07 }        & \textbf{36.97 }    & \textbf{30.96}     & \textbf{33.65} \\
\bottomrule
\end{tabular}
\end{table*}

\section{Experiments}
\label{sec:experiments}

We first provide information necessary for replicating our results, including datasets used and hyperparameter settings (Sec.~\ref{subsec:dataset-and_impl}), and then show main results for video regression (Sec.~\ref{subsec:main-results}).
We show the effectiveness of decoder-side parameter-redistribution for improving the appearance of high resolution frames (Sec.~\ref{subsec:component-analysis}).
Finally, we demonstrate compelling initial results for downstream tasks including  decoding speed, video compression, internal generalization, and video inpainting (Sec.~\ref{subsec:downstream-tasks}).
We offer results for extensive ablation studies in Appendix ~\ref{subsec:ablation-study}.

\subsection{Dataset and Implementation Details}
\label{subsec:dataset-and_impl}

We use the Big Buck Bunny (Bunny)~\cite{bigbuckbunny},
UVG~\cite{mercat2020uvg} and DAVIS~\cite{wang2016mcl} datasets.
Bunny has 132 frames with resolution $720 \times 1280$, and we center-crop  $640 \times 1280$ to get tiny spatial size (\eg $1\times2$) for embedding.
UVG has 7 videos~\footnote{Beauty, Bosphorus, HoneyBee, Jockey, ReadySetGo, ShakeNDry, YachtRide} with size $1080 \times 1920$ at FPS 120 of 5s or 2.5s, and we center-crop $960 \times 1920$.
We also take 10 videos~\footnote{bike-packing, blackswan, bmx-trees,	breakdance, camel, car-round, car-shadow,	cows,	dance-twirl, dog} from the DAVIS validation subset ($1080 \times 1920$, 50-200 frames) and center crop the $960 \times 1920$.
Unless otherwise specified, we use the Adam optimizer, with beta as (0.9, 0.999), weight decay as 0, and learning rate  at 0.001 with cosine learning rate decay. We also use batch size as 2 and L2 loss as reconstruction loss function.
$K_\text{max}$ is set as 5, reduction $r$ is set as 1.2 in Table~\ref{tab:blk-comparison}.
We set stride list as (5,4,4,2,2), (5,4,3,2,2), and (5,4,4,3,2) for video resolutions of $640\times1280$, $480\times960$, and $960\times1920$ respectively. 

For evaluation metrics, we use PSNR and MS-SSIM to evaluate reconstruction quality, bits per pixel (bpp) for compression, and pixels per pixel (ppp) for model compactness. We conduct all experiments in Pytorch with RTX2080ti GPUs, where it takes around $8$s per epoch to train a 130 frame video of size $640 \times 1280$.
We choose HNeRV's size to ensure the PSNR lies between 30-40 for fair video reconstruction.
We provide more experiment details such as architecture details, qualitative results, exact numerical results corresponding to plots, and per-video compression results, in the supplementary material.

\subsection{Main Results}
\label{subsec:main-results}

\begin{figure*}[t!]
    \centering
    \includegraphics[width=.95\textwidth]{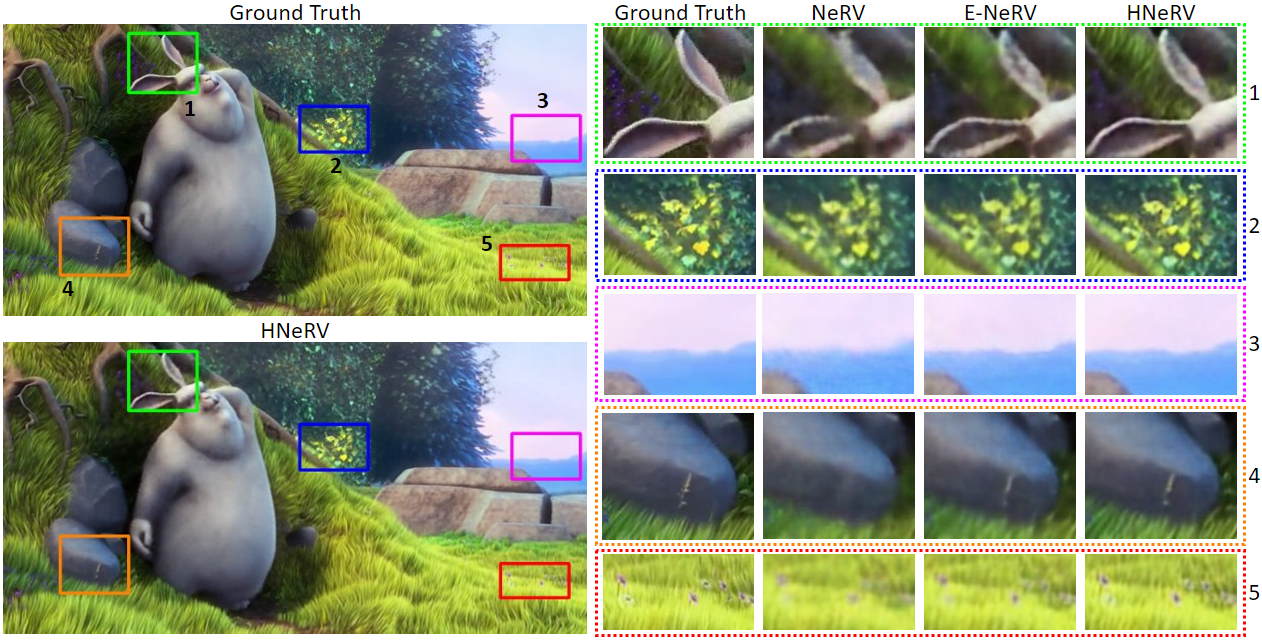}
    \vspace{-1em}
    \caption{
    \textbf{Visualization of video neural representations} at 0.003 ppp, which means the total size is only about 0.3\% of the original video size. 
    On the \textbf{left}, we compare HNeRV to ground truth. On the \textbf{right}, we compare NeRV, E-NeRV, and HNeRV for 5 patches.
    }
    \label{fig:bunny-regression}
    \vspace{-1em}
\end{figure*}

We first compare HNeRV with implicit methods NeRV and E-NeRV on Bunny.
For fair comparison, we scale channel width to make total size comparable as NeRV did.
In Table~\ref{tab:bunny-size}, with the same size and 300 epochs, HNeRV outperforms both NeRV and E-NeRV.
We also show comparison of different training time in Table~\ref{tab:bunny-epoch} with 0.75M size and in Figure~\ref{fig:hnerv-teaser} (right) with 1.5M size, where HNeRV converges much faster compared to implicit methods.
We show improvements qualitatively as well in Figure~\ref{fig:bunny-regression}.
As a compact representation, HNeRV reconstructs the video well with only 0.35M parameters, at 0.003 ppp.
We also evaluate it on 7 UVG videos and 4 DAVIS videos, where HNeRV shows large improvements at resolution  \textbf{$960\times1920$} in Table~\ref{tab:video-960p}, and its resized \textbf{$480\times960$} version in Table~\ref{tab:video-480p}, with size 3M and 300 epochs.

\subsection{Parameter Distribution Analysis}
\label{subsec:component-analysis}

\begin{figure}[t!]
    \centering
    \includegraphics[width=.85\linewidth]{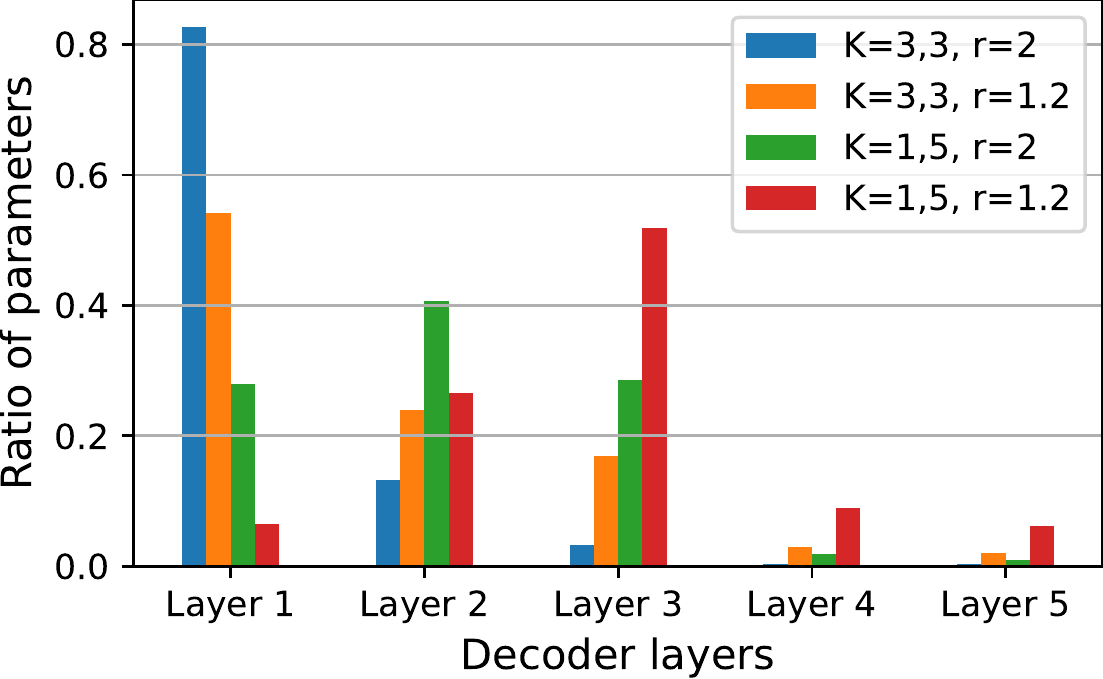}
    \vspace{-0.5em}
    \caption{Parameter distributions for decoder blocks. See \cref{tab:nerv-distributed} for PSNR and MS-SSIM results with these 4 settings.}
    \label{fig:decoder-param-distribution}
    \vspace{-1em}
\end{figure}

As part of our novel architectural innovation, we balance the parameters in the HNeRV decoder.
While NeRV-like architectures naturally have a vanishingly small number of parameters in the final layers, \cref{fig:decoder-param-distribution} shows how we adjust such that the initial and final layers have a roughly equal number of parameters.
Table \ref{tab:nerv-distributed} demonstrates how more even parameter distribution achieves optimal results not only for HNeRV, but also for NeRV\footnote{We label NeRV with $K_\text{max}$=5 as (1,5) since its FC layer can be seen as a $1 \times 1$ convolution layer.}: setting K$_{min}$ to $1$ and K$_{max}$ to $5$, with $r$ at $1.2$, maximizes PSNR and MS-SSIM for both architectures.
\cref{tab:hnerv-ks-ablation} and \cref{tab:hnerv-reduction-ablation} further verify that these hyperparameters are optimal for HNeRV.
We offer these results as strong evidence that our parameter balancing approach not only enables the success of HNeRV, but would be broadly applicable to other NeRV-like architectures.

\begin{table}[t!]
\centering
\caption{Analysis of parameter rebalancing.}
\label{tab:nerv-distributed}
    \vspace{-0.5em}
    \resizebox{.75\linewidth}{!}{
\begin{tabular}{@{}cc|cc|cc@{}}
\toprule
    &     & \multicolumn{2}{c}{NeRV} & \multicolumn{2}{c}{HNeRV} \\
K   & r   & PSNR       & MS-SSIM     & PSNR       & MS-SSIM      \\
\midrule    
3,3 & 2   & 30.87      & 0.9341      & 29.91      & 0.9203       \\
3,3 & 1.2 & 32.27      & 0.9496      & 33.09      & 0.9587       \\
1,5 & 2   & 31.34      & 0.9399      & 34.32      & 0.9715       \\
1,5 & 1.2 & \textbf{33.03}      & \textbf{0.9573}      & \textbf{35.57}      & \textbf{0.9773}      \\
\bottomrule
\end{tabular}
}
\end{table}

\subsection{Downstream Tasks}
\label{subsec:downstream-tasks}

\begin{figure*}[]
    \centering
    \includegraphics[width=.9\textwidth]{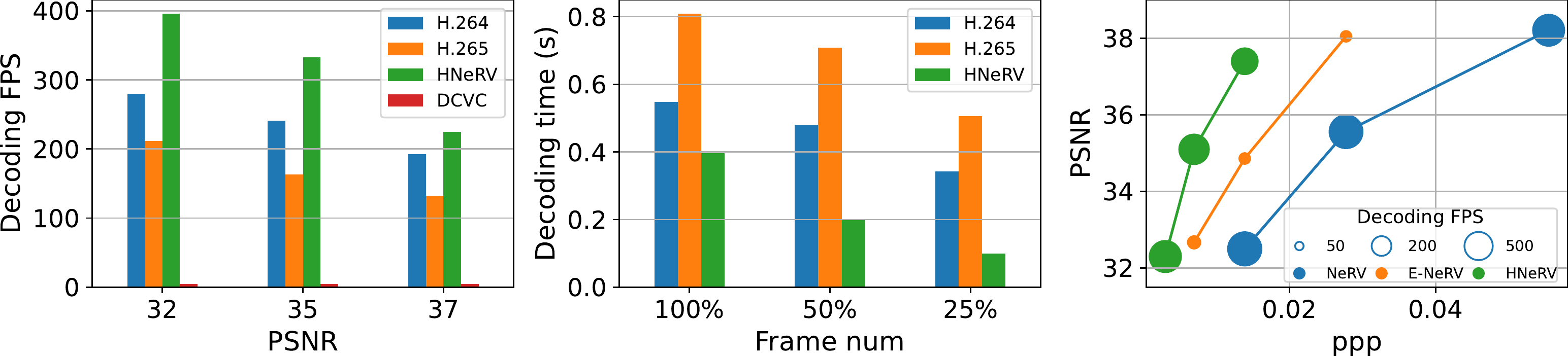}
    \vspace{-.8em}
    \caption{\textbf{Video decoding.}
    \textbf{Left}: HNeRV outperforms traditional video codecs H.264 and H.265, and learning-based compression method DCVC.
    \textbf{Middle}: HNeRV shows much better flexibility when decoding only a portion of video frames.
    \textbf{Right}: HNeRV performs well for compactness (ppp), reconstruction quality (PSNR), and decoding speed (FPS).}    
    \label{fig:decoding-all}

\vspace{2em}
    \centering    
\begin{minipage}{0.5\linewidth}    
    \includegraphics[width=.98\linewidth]{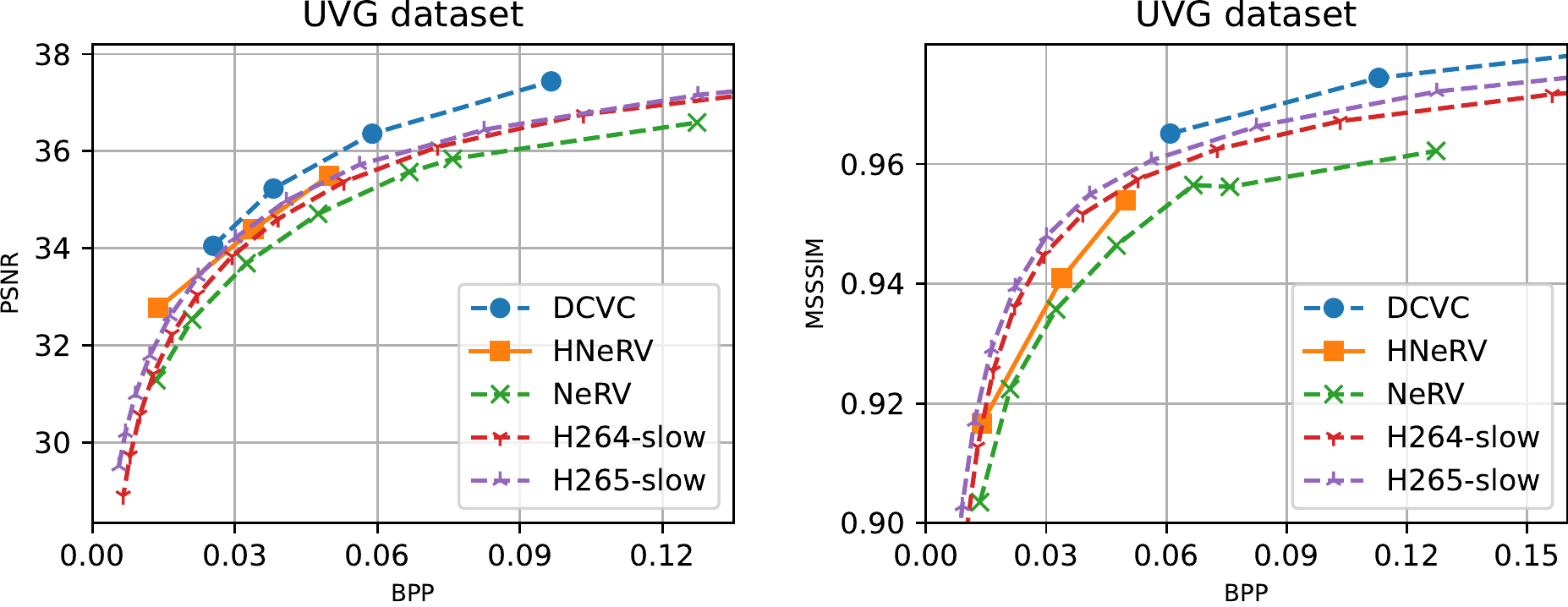}
    \vspace{-0.5em}
    \caption{\textbf{Compression} results on UVG dataset.}
    \label{fig:compress-uvg}
\end{minipage} 
\hfill
\begin{minipage}{0.49\linewidth}   
\includegraphics[width=.98\linewidth]{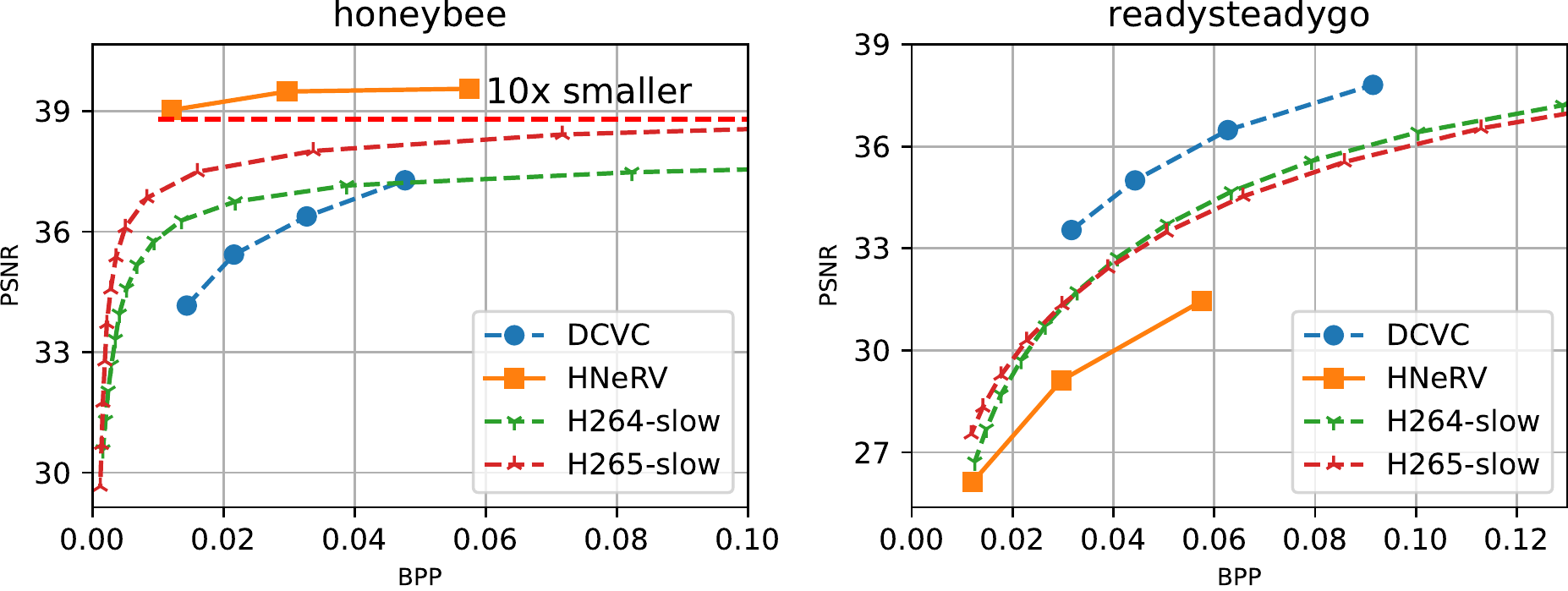}
    \vspace{-0.5em}
    \caption{\textbf{Best/worst} compression cases from UVG dataset. 
    }  
    \label{fig:compress-best-worst}
    \vspace{-1.em}
\end{minipage}     
\end{figure*}

\begin{table*}[t!]
\centering
\caption{\textbf{Internal generalization} results. NeRV, E-NeRV, and HNeRV use interpolated embeddings as input, HNeRV$\dagger$ uses held-out frames as input.
With content-adaptive embeddings as input, HNeRV shows much better reconstruction on held-out frames}
\label{tab:video-interpolation}
    \resizebox{.9\textwidth}{!}{
\begin{tabular}{@{}l|ccccccccccc|c@{}}
\toprule
Method & beauty & swan & bmx & bosph & dance & camel & bee & jockey & ready & shake & yach & avg. \\
\midrule
NeRV    & 28.05  & 17.94          & 15.55          & 30.04    & 16.99           & 14.83      & 36.99    & 20.00  & 17.02         & 29.15     & 24.50     & 22.82 \\
HNeRV & \underline{30.97} &	\underline{21.44}	& \underline{17.35} &	\underline{34.38} &	\underline{20.2}	& \underline{19.93} &	\underline{38.83} &	\underline{23.67} &	\underline{20.90} &	\textbf{32.69} &	\textbf{27.30} &	\underline{26.15} \\
HNeRV$\dagger$   & \textbf{31.10}  & \textbf{21.97}          & \textbf{18.29}          & \textbf{34.38}    & \textbf{20.29}           & \textbf{20.64 }     & \textbf{38.83}    & \textbf{23.82}  & \textbf{20.99}         & \underline{}{32.61}     & \underline{27.24}     & \textbf{26.38} \\
\bottomrule
\end{tabular}
}        

\centering
\caption{Video \textbf{inpainting} results (PSNR $\uparrow$) with 5 fixed box masks on input videos. `Input' is the baseline of mask video and ground truth}
\label{tab:video-inpainting}
\vspace{0.6em}
    \resizebox{.9\textwidth}{!}{
\begin{tabular}{@{}l|cccccccccc|c@{}}
\toprule
 Video     & bike & b-swan & bmx & b-dance & camel & c-round & c-shadow & cows  & dance-twirl & dog   & avg.   \\
\midrule      
Input & 23.14        & 20.24     & 19.99     & 21.36      & 17.3  & 20.47          & 18.92      & 19.37 & 20.45       & 18.39 & 19.96 \\
NeRV  & 30.94        & 33.43     & 32.07     & 27.82      & 31.99 & 29.09          & 31.63      & 30.08 & 30.45       & 33.85 & 31.14 \\
IIVI & \textbf{31.87} &	\textbf{36.02} &	\textbf{34.36} &	\underline{27.63} &	\textbf{35.11} &	\textbf{32.61} &	\textbf{33.69} &	\textbf{31.26} &	\textbf{31.44} &	\textbf{35.7} &	\textbf{32.97} \\
HNeRV & \underline{31.27} &	\underline{34.24} &	\underline{33.95} &	\textbf{27.94} &	\underline{32.21} &	\underline{30.88} &	\underline{33.07} &	\underline{30.82} &	\underline{31.21} &	\underline{34.7} &	\underline{32.03} \\
\bottomrule
\end{tabular}
}
\end{table*}

\noindent\textbf{Video decoding.}
We evaluate video decoding on Bunny with channel reduction $r$ as 1.5, where H.264 and H.265 are tested with 4 CPUs\footnote{Intel(R) Xeon(R) Silver 4216 CPU @ 2.10GHz}, while DCVC~\cite{li2021dcvc} and HNeRV are tested with 1 GPU~(RTX2080ti).
We only measure the forward time for DCVC and HNeRV.
We compare video decoding at various reconstruction qualities (PSNR at  32, 35, and 37) in Figure~\ref{fig:decoding-all} (left), where HNeRV outperforms traditional codecs (H.264 and H.265) and learning-based DCVC.
Note that although many prior learning-based compression methods show bit-distortion improvements, their decoding speeds \textit{lag far behind} traditional codecs and neural representation.
Besides, most compression methods encode and decode frames in an auto-regressive way and can not access frames randomly.
Compared to these methods, the decoding of HNeRV is much simpler and can be deployed to any platform easily.
We compare decoding time in Figure~\ref{fig:decoding-all} (Middle) (PSNR at 35) where $100\%, 50\%$, and $25\%$ of frames (evenly sampled, to mimic \eg a discrete reduction of FPS) are decoded.
Since there is no dependency among video frames, HNeRV can decode them in parallel and decoding time decreases linearly with the number of frames decoded.
In contrast, H.264 and H.265 still need to decode most frames, even though only some of them are needed.
Finally, we compare with implicit methods in Figure~\ref{fig:decoding-all} (right), where HNeRV is slightly slower than NeRV since the computation of later layers is more expensive due to large $K$ and channel width.
As a hybrid neural representation, HNeRV achieves much better trade-offs with respect to compactness (ppp), reconstruction quality (PNSR), and decoding speed (FPS).

\noindent\textbf{Video compression.}
With model pruning (10\% pruned), embedding quantization (8 bits), model quantization (8 bits), and model entropy encoding (8\% saved), we show video compression results on UVG in Figure~\ref{fig:compress-uvg}.
HNeRV outperforms the implicit method, NeRV, and traditional video codecs H.264 and H.265.
\textit{Note that HNeRV achieves this using a small model for each video, while NeRV fits a big model on concat videos (for better compression) which greatly slows down the encoding and decoding speed.}
We also show the best and worst cases of HNeRV video compression for the `honeybee' and `readysteadygo' videos respectively in Figure~\ref{fig:compress-best-worst}, where HNeRV achieves outstanding performance when the camera is not moving, such as with the `honeybee' video ($10\times$ smaller than H.265 for equivalent PSNR).
Given the limited performance on videos of highly dynamic scenes, we propose finding a good size and network architecture for such videos as future work.

\begin{figure*}[t]
    \centering    
    \includegraphics[width=.85\textwidth]{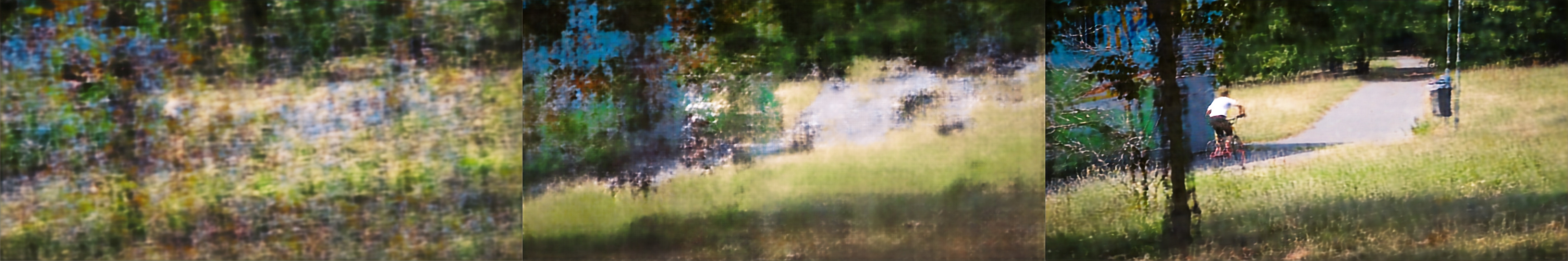}
    \includegraphics[width=.85\textwidth]{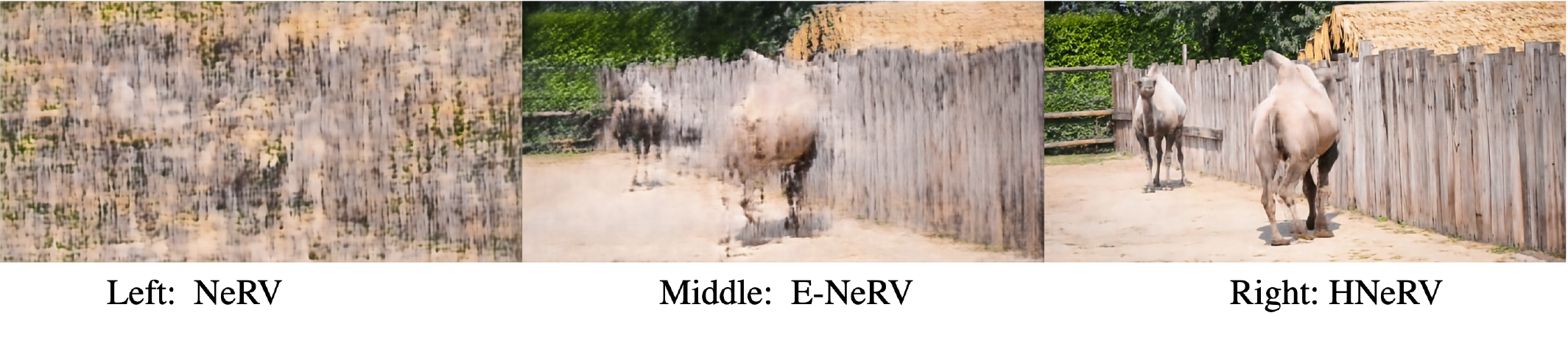}
    \vspace{-1.2em}
    \caption{Visualization of \textbf{Embedding interpolation}.}
    \label{fig:interpolate}    

    \vspace{1.5em}
    \centering
    \includegraphics[width=.85\linewidth]{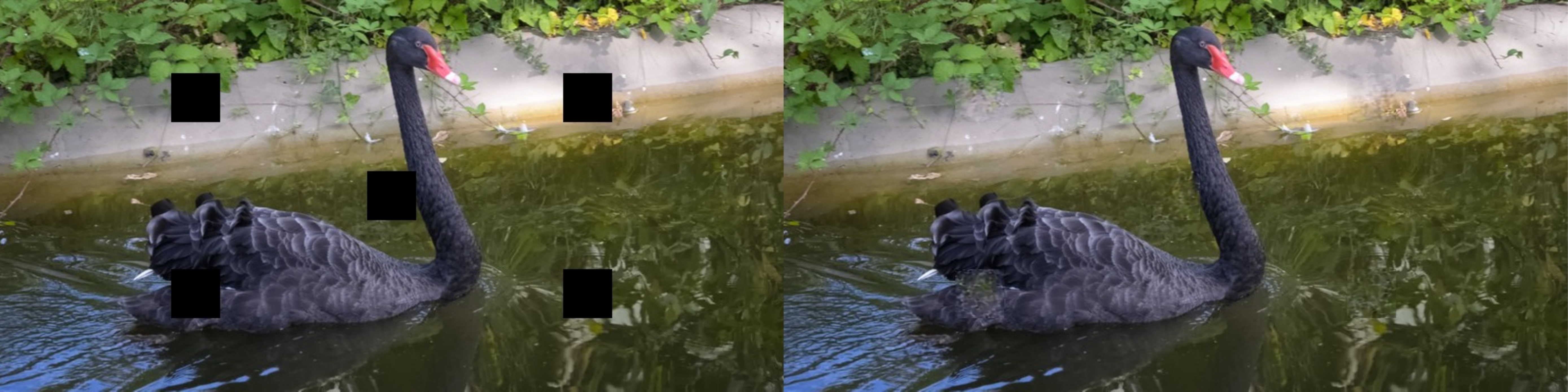}
    \includegraphics[width=.85\linewidth]{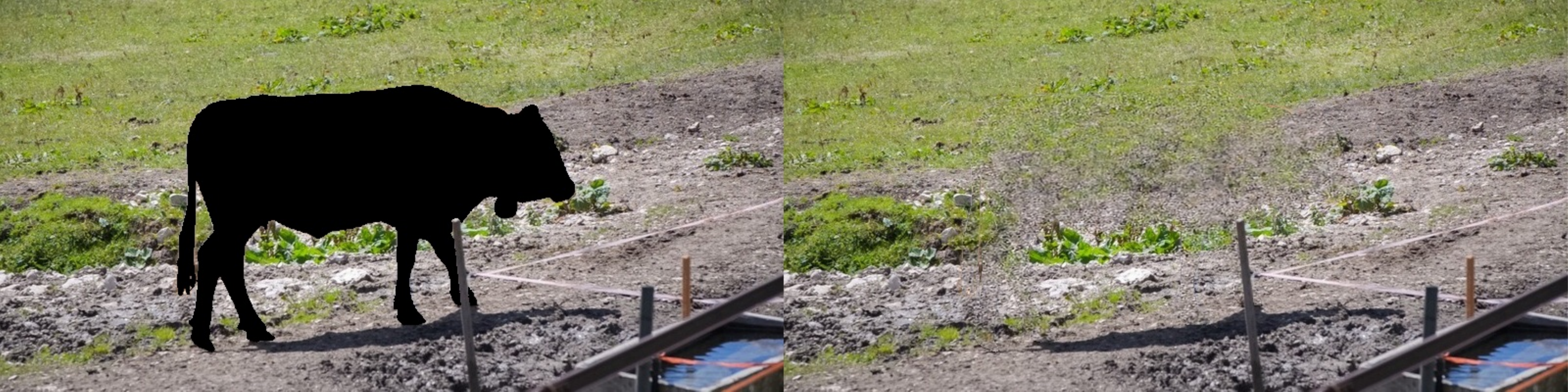} 
    \vspace{-0.5em}
    \caption{\textbf{Inpainting} results of fixed masks and object masks.
    \textbf{Left)} input frame; \textbf{Right)} HNeRV output.}    
    \label{fig:inpaint}    
\end{figure*}

\noindent\textbf{Internal generalization.}
Since HNeRV leverages content-adaptive embeddings, we also evaluate it for the video interpolation task.
Holding out every other frame as a test set, NeRV, E-NeRV, and HNeRV use interpolated embedding as input, while HNeRV$\dagger$ uses the test frame as input.
With learnable and content-adaptive embedding, our HNeRV shows much better generalization, quantitatively in Table~\ref{tab:video-interpolation} and qualitatively in Figure~\ref{fig:interpolate}.

\noindent\textbf{Video inpainting.}
We also explore video inpainting with fixed and object masks.
For fixed masks, we use 5 boxes of width 50 (Figure~\ref{fig:inpaint} (top)) and show quantitative results in Table~\ref{tab:video-inpainting} where HNeRV improves inpainting performance over the implicit method, NeRV.
Although we do not have any specific design for the inpainting task, HNeRV even achieves comparable performance with an SOTA inpainting method, IIVI~\cite{ouyang2021video}.
We show qualitative results in Figure~\ref{fig:inpaint}.

\section{Conclusion}
\label{sec:conclusion}

In this paper, we propose a hybrid neural representation for videos~(HNeRV).
With content-adaptive embedding and evenly-distributed parameters, HNeRV improves video regression performance compared to implicit methods in terms of reconstruction quality, convergence speed, and internal generalization.
As a video representation, HNeRV is also simple, fast, and flexible for video decoding, and shows good performance for video compression and inpainting.

There are many limitations of HNeRV as well.
Firstly, as a neural representation, HNeRV stores each video as a neural network.
Given a new video, HNeRV still needs time to train to fit the video.
Secondly, although HNeRV can represent a video well, finding a best-fit embedding size, model size, and network architecture (K$_\text{max}$, $r$, \etc) remains an open problem.
Finally, although increasing kernel sizes and channel widths at later layers largely improves the regression performance, it slightly slows down the network, as shown in Figure~\ref{fig:decoding-all} (right).

\noindent\textbf{Acknowledgements.} This project was partially funded by the DARPA SAIL-ON (W911NF2020009) program, an independent grant from Facebook AI, and Amazon Research Award to AS.

\clearpage
{\small
\bibliographystyle{ieee_fullname}
\bibliography{egbib}
}

\clearpage
\appendix

\setcounter{page}{1}

\twocolumn[
{
\centering
\Large
\textbf{HNeRV: A Hybrid Neural Representation for Videos} \\ 
     {\small Supplementary Material} \\
\vspace{2.0em}
} 
]
\section{Ablation study}
\label{subsec:ablation-study}

We show the effectiveness of even-distributed parameters in Table~\ref{tab:hnerv-ks-ablation} and Table~\ref{tab:hnerv-reduction-ablation} by increasing kernel size and channel width of later layers.
For the NeRV block, it uses fixed $K = 3$, and channel reduction factor $r = 2$.
We also show an embeddings ablation study, for spatial size~($h \times w$) in Table~\ref{tab:spatial-size-ablation} and embedding dimensions ($d$) in Table~\ref{tab:embed-dim-ablation}.

\begin{table}[h!]
\begin{minipage}{0.24\textwidth}
\caption{\textbf{Kernel size} $(K_\text{min}$, $K_\text{max}$) ablation, (with r=1.2)}
\label{tab:hnerv-ks-ablation}
\vspace{-1em}
    \resizebox{.98\textwidth}{!}{
    \setlength{\tabcolsep}{10pt}
\begin{tabular}{@{}l|cc@{}}
\toprule
$K$ & PSNR & MS-SSIM  \\
\midrule
1,3 &	35.02 &	0.9752 \\
1,5 &	\textbf{35.57} & \textbf{0.9773} \\
1,7 &	35.07 &	0.9757    \\
3,3 &  33.09 &	0.9587 \\
\bottomrule
\end{tabular}
}
\end{minipage} 
\hfill
\begin{minipage}{0.23\textwidth}
\caption{\textbf{Channel reduction $r$} ablation, (with K=1,5)}
\label{tab:hnerv-reduction-ablation}
\vspace{-1em}
    \resizebox{.98\textwidth}{!}{
    \setlength{\tabcolsep}{8pt}
\begin{tabular}{@{}l|cc@{}}
\toprule
$r$ & PSNR & MS-SSIM  \\
\midrule
1 &	34.96	& 0.9745 \\
1.2 &	\textbf{35.57} &	\textbf{0.9773} \\
1.5 &	34.98 &	0.9762    \\
2 &  34.32	& 0.9715 \\
\bottomrule
\end{tabular}
}
\end{minipage} 
\end{table}

\begin{table}[h!]
    \centering
\begin{minipage}{0.24\textwidth}
\caption{\textbf{Embedding spatial size} ablation}
\label{tab:spatial-size-ablation}
\vspace{-0.5em}
    \resizebox{.98\textwidth}{!}{
    \setlength{\tabcolsep}{5pt}
\begin{tabular}{@{}l|cc@{}}
\toprule
$h \times w$  & PSNR & MS-SSIM  \\
\midrule
$1\times2$ &	34.79 &	0.9735 \\
$2\times4$	& \textbf{35.57} &	\textbf{0.9773} \\
$4\times8$	& 35.12 &	0.9761 \\
\bottomrule
\end{tabular}
}
\end{minipage} 
\hfill
\begin{minipage}{0.23\textwidth}
\caption{\textbf{Embedding dimension} ablation}
\label{tab:embed-dim-ablation}
\vspace{-0.5em}
    \resizebox{.98\textwidth}{!}{
    \setlength{\tabcolsep}{8pt}
\begin{tabular}{@{}l|cc@{}}
\toprule
$d$ & PSNR & MS-SSIM  \\
\midrule
8	& 35.13 &	0.9770 \\
16	&\textbf{35.57} &	\textbf{0.9773} \\
32 &	35.08	& 0.9758 \\
\bottomrule
\end{tabular}
}
\end{minipage} 
\end{table}

\section{Video decoding}
We firstly show command to evaluate decoding speed of H.264 and H.265: \\
\quad \quad ffmpeg -threads ThreadsNum  -i Video -preset medium -f null -benchmark -

And we also show quantitative decoding results in Table~\ref{append-tab:decode-speed-main}, \ref{append-tab:decode-speed-frames}, and Table~\ref{append-tab:decode-speed-reduction-ablation}.
In Table~\ref{append-tab:decode-speed-reduction-ablation}, we can further increase video decoding speed with a smaller channel width (\ie a big reduction factor $r=2$).

\begin{table}[h!]
\centering
\begin{minipage}{0.49\linewidth} \centering
\centering
\caption{Decoding FPS $\uparrow$ }
\label{append-tab:decode-speed-main}
\resizebox{.98\textwidth}{!}{
\renewcommand{\arraystretch}{1.1}
\setlength{\tabcolsep}{5pt}
    \begin{tabular}{@{}l|ccc@{}}
    \toprule
PSNR &	32	& 35 &	37 \\
\midrule
H.264 &	279.7 &	240.9 &	192.7 \\
H.265 & 211.9 &	163.2 &	132.5 \\
DCVC & 4.7	& 4.6 &	4.5 \\
HNeRV & \textbf{395.9}	& \textbf{332.7}	& \textbf{224.8} \\
    \bottomrule
    \end{tabular}
}
\end{minipage} 
\hfill
\begin{minipage}{0.49\linewidth} \centering
\caption{Decoding time (s) $\downarrow$} 
\label{append-tab:decode-speed-frames}
\resizebox{.98\textwidth}{!}{
\renewcommand{\arraystretch}{1.2}
\setlength{\tabcolsep}{5pt}
    \begin{tabular}{@{}l|ccc@{}}
    \toprule
\# Frames  &	100\% &	50\% 	& 25\%  \\
\midrule
H.264	& 0.548	& 0.480	& 0.343 \\
H.265	& 0.809	& 0.708	& 0.506 \\
DCVC	& 27.913 &	24.424 &	17.446 \\
HNeRV & \textbf{0.397}	& \textbf{0.198}	& \textbf{0.099} \\
    \bottomrule
    \end{tabular}
}
\end{minipage} 
\begin{minipage}{0.55\linewidth} \centering
\centering
\vspace{1.5em}
\caption{HNeRV Decoding FPS }
\label{append-tab:decode-speed-reduction-ablation}
\resizebox{.98\textwidth}{!}{
\renewcommand{\arraystretch}{1.1}
\setlength{\tabcolsep}{5pt}
    \begin{tabular}{@{}l|ccc@{}}
    \toprule
PSNR &	32	& 35 &	37 \\
\midrule
r=1.5 & 	395.9 &	332.7 &	224.8 \\
r=1.75 &	397.4 &	373.8 &	320.7 \\
r=2 &	    \textbf{405.5} &	\textbf{383.3} &	\textbf{350.5} \\
    \bottomrule
    \end{tabular}
}
\end{minipage} 
\end{table}

\section{Video compression}
Then we show the details for downstream tasks of video compression, which can be divided into three steps: global unstructure pruning, quantization, and entropy encoding.

\textit{1) Model Pruning.}
Given a pre-trained model, we use global unstructured pruning to reduce the model size, where parameters below a threshold are pruned and set as zero.
For a model parameter $\theta_i$,
$
    \theta_i = 
    \begin{cases}
    \theta_i, & \text{if } \theta_i \geq \theta_q\\
    0,              & \text{otherwise,}
    \end{cases}
$    
where $\theta_q$ is the $q$ percentile value for all model parameters $\theta$. As a normal practice, we fine-tune the model to regain the representation after pruning.

\textit{2) Model and embedding quantization.}
Model quantization and embedding quantization follow the same scheme.
Given an vector $\mu$, we linearly map every element to the closest integer,
\begin{equation}
  \begin{aligned}
    \mu_i &= \text{Round}\left(\frac{\mu_i - \mu_\text{min}}{\text{scale}}\right) * \text{scale} + \mu_\text{min} ,
    \text{where } \\
    \text{scale} &= \frac{\mu_\text{max} - \mu_\text{min}}{2^\text{b} - 1}
    \label{equa:quant}
  \end{aligned}
\end{equation}
$\mu_{i}$ is one vector element, `Round' is a function that rounds to the closest integer, `b' is the bit length for quantization, $\mu_\text{max}$ and $\mu_\text{min}$ are the max and min value of vector $\mu$, and `scale' is the scaling factor.
For scaling factor and zero points at this step, we can also try other methods instead of current min-max one, like choosing $2^b$ evenly-distributed values to minimum the mean square error.

\textit{3) Entropy encoding.}
Finally, we use entropy encoding to further reduce the size.
Specifically, we leverage Huffman coding~\cite{huffman1952method} for quantized weights and get lossless compression.

\section{Weight Pruning for Model Compression.}
We appreciate this concern, which has been unresolved since the original NeRV paper.
By applying entropy encoding (assigning fewer bits for frequent symbols), we can store pruned weights with limited bits, since all pruned weights share a frequent symbol: $0$.
We provide corrected model compression results in  \cref{tab:compression-entropy}, and will update the paper accordingly.
We use 2 baselines (models with no pruning) -- one where the model is only quantized, and another where we apply entropy encoding after quantization.
As we prune more parameters, entropy encoding enables us to use fewer bits to store the sparse model weights.

\begin{table}[h]
\centering
\vspace{-1em}
\caption{Compression results. 
``Size ratio'' compares to model with quant. only, and ``Sparsity'' indicates amount of weights pruned. }
\label{tab:compression-entropy}
\vspace{-0.8em}
\resizebox{.98\linewidth}{!}{
    \begin{tabular}{@{}l|c|ccccc@{}}
    \toprule
  Compression  & Quant & \multicolumn{5}{c}{Prune + Quant + Entropy coding} \\
  \hline
Sparsity          & 0\% & 0\%  & 10\%   & 15\%   & 20\%   & 25\%   \\

    \midrule
PSNR                 & 37.61 & 37.56  & 37.51  & 37.32  & 37.02  & 36.61  \\
Size (bits) & 11.54M & 10.94M  & 10.41M  & 10.09M  & 9.77M   & 9.36M     \\
Size ratio & 100\% & 94.8\% & 90.2\% & 87.4\% & 84.7\% & 81.1\% \\

    \bottomrule
    \end{tabular}
}
\end{table}

\section{HNeRV architecture details}
We also provide architecture details for HNeRV models in various tasks and datasets in 
Table~\ref{append-tab:archi-details}, with total size, strides list, encoder dimension $c1$, embedding dimension $d$, channel width of decoder input $c2$, channel reduction $r$, lowest channel width $Ch_\text{min}$, min and max kernel size K$_\text{min}$, K$_\text{max}$
.
\begin{table}[h!]
\centering
\footnotesize
\caption{HNeRV architecture details }
\label{append-tab:archi-details}
\resizebox{.98\linewidth}{!}{
    \begin{tabular}{@{}l|cccccccc@{}}
    \toprule
Video size & size & strides & c1 & d & c2 & r & Ch$_\text{min}$ & K$_\text{min}$, K$_\text{max}$ \\
\midrule
640$\times$1280 & 0.35 & 5,4,4,2,2 & 64 & 16 & 32 & 1.2 & 12 & 1,5 \\
640$\times$1280 & 0.75 & 5,4,4,2,2 & 64 & 16 & 48 & 1.2 & 12 & 1,5 \\
640$\times$1280 & 1.5 & 5,4,4,2,2 & 64 & 16 & 68 & 1.2 & 12 & 1,5 \\
640$\times$1280 & 3 & 5,4,4,2,2 & 64 & 16 & 97 & 1.2 & 12 & 1,5 \\
480$\times$960 & 3 & 5,4,3,2,2 & 64 & 16 & 110 & 1.2 & 12 & 1,5 \\
960$\times$1920 & 3 & 5,4,4,3,2 & 64 & 16 & 92 & 1.2 & 12 & 1,5 \\
    \bottomrule
    \end{tabular}
}
\end{table}

\section{Per-video compression results}
\begin{figure*}
    \centering
      \includegraphics[width=.8\textwidth]{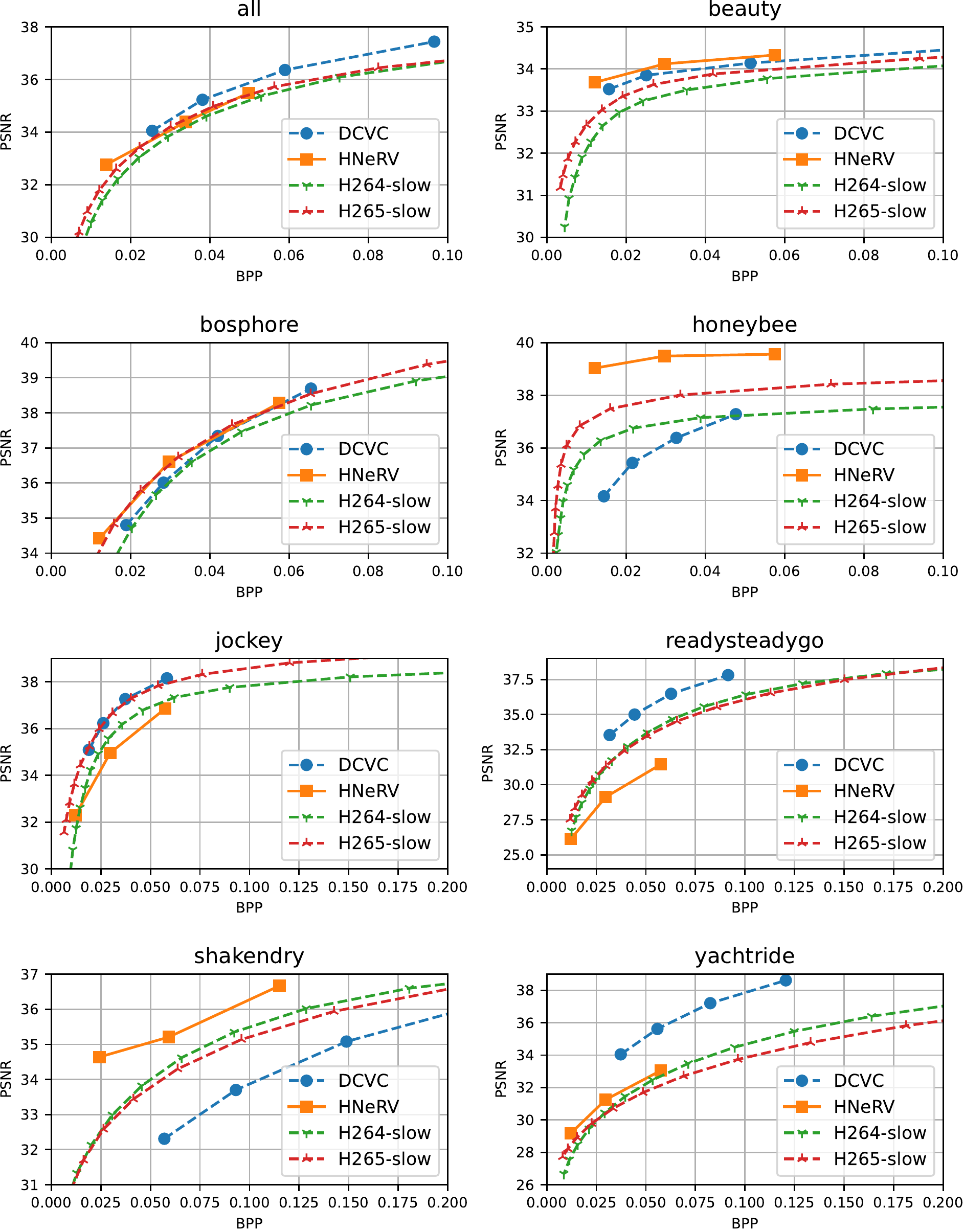}
      \caption{\textbf{Compression} results averaged across all \textbf{UVG} videos, and for each specific videos.}  
    \label{append-fig:all-videos-compression}
\end{figure*}

We also show video compression results for \textbf{UVG} videos in Figure~\ref{append-fig:all-videos-compression}.

\section{More visualizations}
\begin{figure*}[h!]
    \centering    
    \includegraphics[width=.98\textwidth]{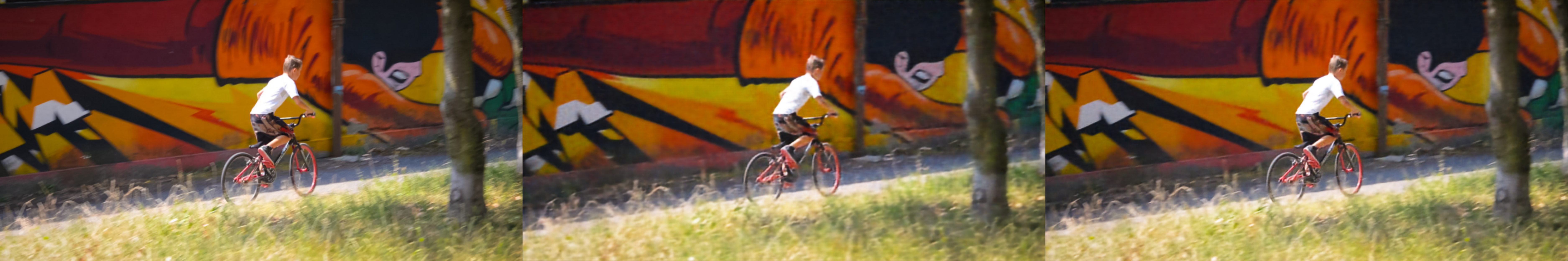}
    \includegraphics[width=.98\textwidth]{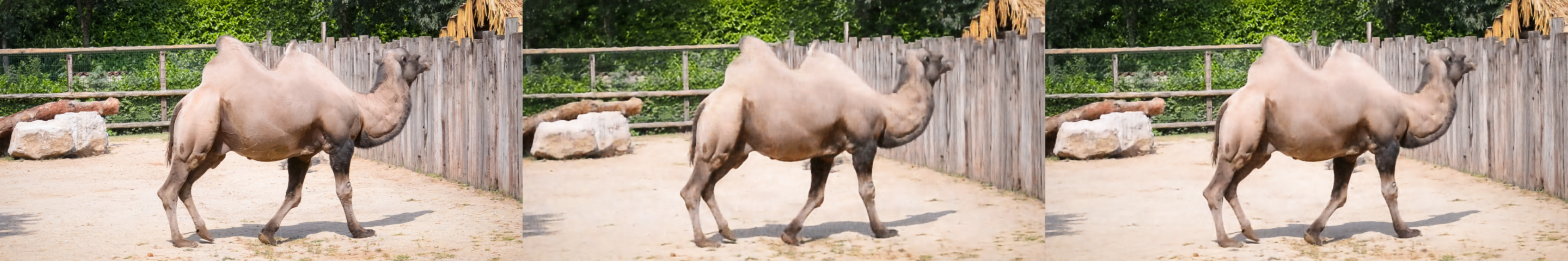}
    \includegraphics[width=.98\textwidth]{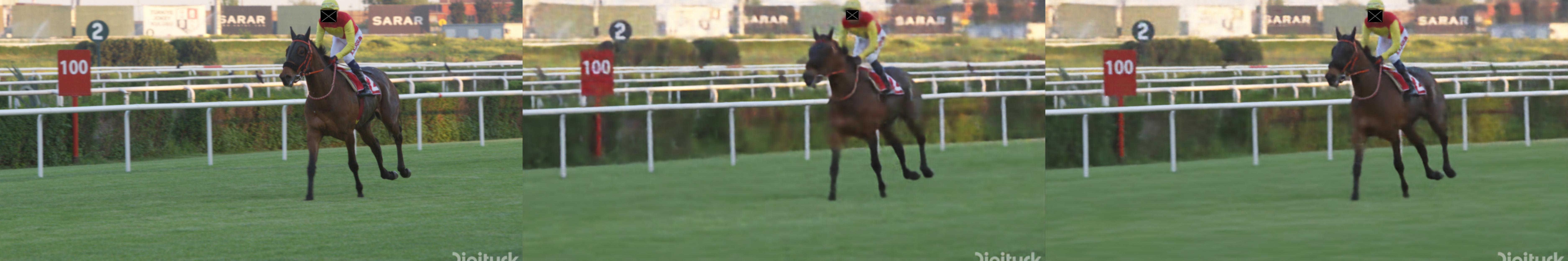}
    \includegraphics[width=.98\textwidth]{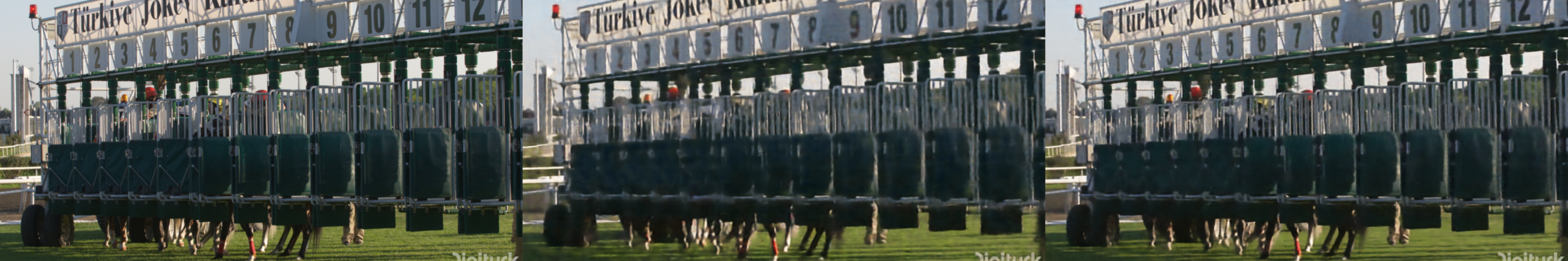}
    \caption{\textbf{Video regression} results. 
    \textbf{Left)} ground truth.
    \textbf{Middle)} NeRV output.
    \textbf{Right)} HNeRV output.}
    \label{append-fig:regression}
\end{figure*}

\begin{figure*}[h!]
    \centering
    \includegraphics[width=.98\textwidth]{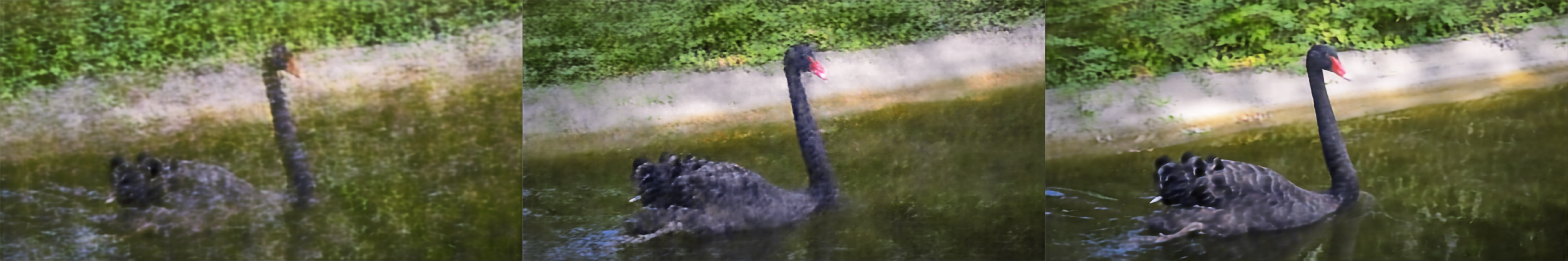}
    \caption{\textbf{Interpolation} results.}
    \label{append-fig:interpolate}
\end{figure*}

\begin{figure*}[h!]
    \centering    ----------
    \includegraphics[width=.98\textwidth]{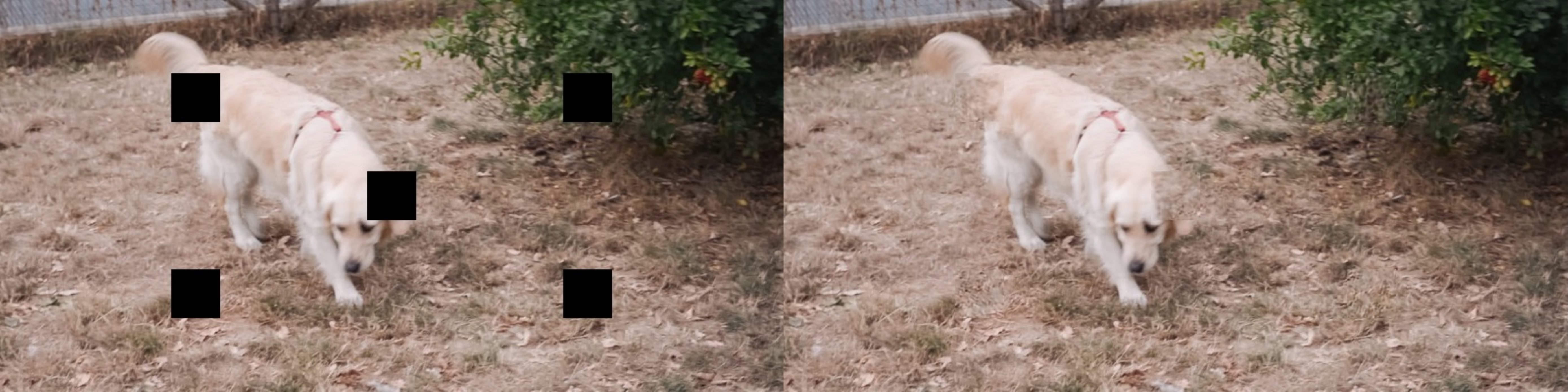}
    \includegraphics[width=.98\textwidth]{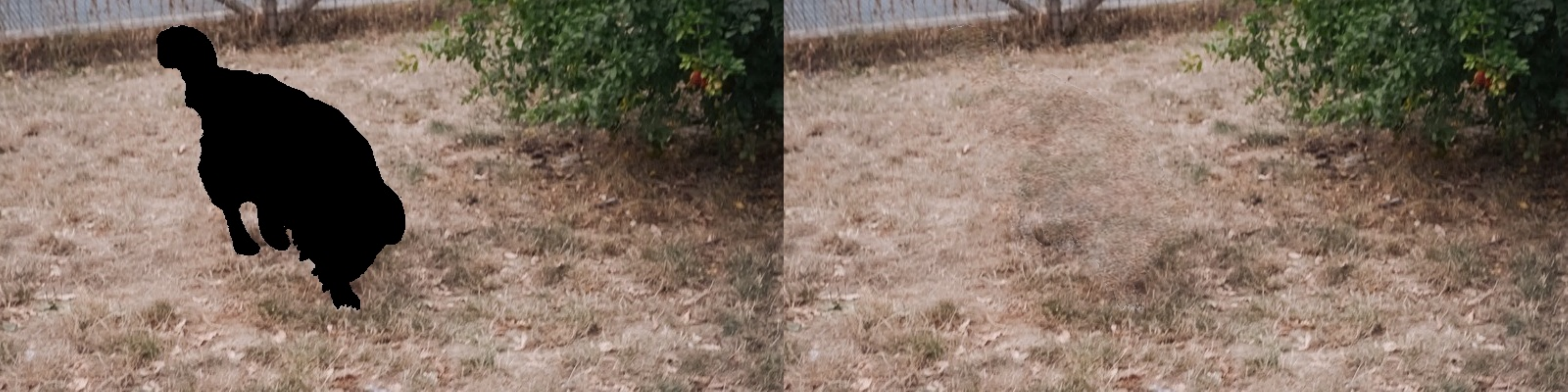}  
    \caption{\textbf{Inpainting} results.}
    \label{append-fig:inpaint}
\end{figure*}

We show more visualizations for video regression (Figure~\ref{append-fig:regression}), video interpolation (Figure~\ref{append-fig:interpolate}), and video inpainting (Figure~\ref{append-fig:inpaint}).

\end{document}